\documentclass[ejs,preprint]{imsart}

\RequirePackage[OT1]{fontenc}
\RequirePackage{amsthm,amsmath}
\RequirePackage[authoryear]{natbib}
\RequirePackage[colorlinks,citecolor=blue,urlcolor=blue]{hyperref}
\usepackage{graphicx}

\usepackage{cleveref}
\usepackage{tikz}
\usepackage[ruled,vlined,linesnumbered]{algorithm2e}
\usepackage{caption}
\usepackage{subcaption}
\usepackage[top=3cm, bottom=3cm, left=2.5cm, right=2.5cm]{geometry}

\usepackage{booktabs}

\usepackage{color}
\usepackage{latexsym}              
\usepackage{amssymb}               
\usepackage{amsfonts}              

\usepackage{multirow}
\usepackage{tikz}                  
\usetikzlibrary{arrows,positioning,shapes}
\usepackage{tcolorbox}

\startlocaldefs
\numberwithin{equation}{section}
\theoremstyle{plain}
\newtheorem{theorem}{Theorem}[section]
\newtheorem{definition}{Definition}[section]
\newtheorem{proposition}{Proposition}[section]
\newtheorem{remark}{Remark}[section]
\newtheorem{lemma}{Lemma}[section]
\newtheorem{corollary}{Corollary}[section]
\endlocaldefs

\newcommand{\KL}{\operatorname{KL}}

\begin{document}

\begin{frontmatter}
\title{Upper and Lower Bounds on the Performance of Kernel PCA}
\runtitle{Upper and Lower Bounds on the Performance of Kernel PCA}

\begin{aug}
\author{\fnms{Maxime} \snm{Haddouche}\thanksref{t2,t3}\ead[label=e1]{maxime.haddouche@inria.fr}}
\author{\fnms{Benjamin} \snm{Guedj}\thanksref{t2,t3}\ead[label=e2]{benjamin.guedj@inria.fr}}
\and \author{\fnms{John} \snm{Shawe-Taylor}\thanksref{t3}\ead[label=e4]{j.shawe-taylor@ucl.ac.uk}}

\address{
\printead{e1,e2,e4}}

\thankstext{t2}{Inria, France}
\thankstext{t3}{University College London, United Kingdom}
\runauthor{Haddouche, Guedj and Shawe-Taylor}

\end{aug}

\begin{abstract}
  Principal Component Analysis (PCA) is a popular method for dimension reduction and has attracted an unfailing interest for decades. More recently, kernel PCA (KPCA) has emerged as an extension of PCA but, despite its use in practice, a sound theoretical understanding of KPCA is missing. We contribute several lower and upper bounds on the efficiency of KPCA, involving the empirical eigenvalues of the kernel Gram matrix and new quantities involving a notion of variance. These bounds show how much information is captured by KPCA on average and contribute a better theoretical understanding of its efficiency. We demonstrate that fast convergence rates are achievable for a widely used class of kernels and we highlight the importance of some desirable properties of datasets to ensure KPCA efficiency.
\end{abstract}


\tableofcontents
\end{frontmatter}



\section{Introduction}
Principal Component Analysis (PCA) is a celebrated dimension reduction method. It was first described by \citep{pearson1901}, and it was developed further by several authors \cite[see \emph{e.g.}][and references therein]{jolliffe1986}. In a nutshell, PCA summarises high-dimensional data $(x_1,...,x_m)\in\mathbb{R}^d$ (where $m\in\mathbb{N}^*$), into a smaller space, which is designed to be `meaningful' and more easily interpretable. By `meaningful' we mean that this new subspace still captures efficiently the correlations between data points, while at the same time reducing drastically the dimension of the space. A popular tool to design this meaningful subspace is the \emph{Gram matrix} of the data, defined as $(\langle x_i,x_j\rangle)_{i,j}$. PCA then considers the eigenvectors of this matrix. Note that this is a linear operation, in the sense that PCA consists of an orthogonal transformation of the coordinate system in which we describe our data, followed by a projection onto the first $k$ directions in the new system, corresponding to the largest $k$ eigenvalues of the Gram matrix.

Over the past two decades, PCA has been studied and enriched \citep[\emph{e.g.}, principal curves as a nonlinear extension of PCA, as done by][]{li2021sequential}. The particular extension of PCA that we focus on is 'kernel PCA' (KPCA, \citealp[which may be traced back to][]{scholkopf1998}). Using a kernel, we map our data into a reproducing kernel Hilbert space\footnote{We refer the reader to  \citet{hein2004kernels} and \citet{hofmann2005tutorial} for an introduction to RKHS.} (RKHS).
The linear PCA then operates in this Hilbert space to yield a finite-dimensional subspace onto which we project new data points. The final step is to assess how close from the original data is this projection. Kernel PCA is popular among practitioners in statistics and machine learning. An example is its widespread use in machine learning \citep[\emph{e.g.},][to name but a few recent works]{kim2020,xu2019,vo2016}, to traffic signs recognition \citep{zhang2018effective}, to fault detection in chemistry and biology \citep[e.g.]{harkat2020machine, yao2018fault} and even Earth observation analysis to tackle climate change \citep{bueso2020nonlinear}. This broad range of kernel PCA users makes the need of a better theoretical understanding even more pressing.

A first theoretical study has been proposed by \citet{shawe-taylor2005eigenspectrum} who derived PAC (Probably Approximately Correct) guarantees for kernel PCA. The PAC bounds proposed by \citet{shawe-taylor2005eigenspectrum} were set up to control the averaged projection of new data points onto a finite-dimensional subspace of the RKHS into which data is embedded. This study has been extended by \citet{blanchard2007statistical} with localised fast rates convergence bounds: they prove upper bounds for the
excess risk of the reconstruction error which give different rates in sample size and dimensionality depending on spectral properties of the covariance operator and thus explore the richness of this classical statistical method. The theoretical study of KPCA was further done by \citet{reiss2020nonasymptotic}: the authors provide a theoretical bound for the reconstruction gap of KPCA which informs us how far our data-dependent projection is from the best possible projection for a fixed dimension. Let us also mention \citet{milbradt2020high} who provided the first oracle bound for the reconstruction error of KPCA.

From a technical point of view, those previous works used tools such as Rademacher complexity \citep{shawe-taylor2005eigenspectrum}, empirical process theory\footnote{See \citet{bartlett2005local}.} \citep{blanchard2007statistical}, concentration inequalities\footnote{See \citet{koltchinskii2017concentration}.} \citep{reiss2020nonasymptotic}.

In this work, we investigate a different route and introduce the first PAC-Bayesian study of kernel PCA which, as a byproduct, allows to replace the Rademacher term in \citet{shawe-taylor2005eigenspectrum} by a Kullback-Leibler divergence (which has a closed form when distributions are Gaussian, and can be approximated by Monte Carlo in other cases). PAC-Bayes theory is a powerful framework to study generalisation properties of randomised predictors, and was introduced in the seminal works of \citet{shawe1997pac,McAllester1998,McAllester1999}. PAC-Bayes theory has then been developed further by \citet{seeger2002,McAllester2003,maurer2004note,catoni2007} among others.
We refer to the recent surveys \citet{guedj2019primer,alquier2021user} and tutorial \citet{GueSTICML}, for details on PAC-Bayes theory.

\paragraph{Our contributions.} Our work provides fully computable guarantees for KPCA efficiency. We furnish a first bound which improve on one from \citet{shawe-taylor2005eigenspectrum} in the sense that fast rates are achievable when the dimension of the projected subspace is sufficiently high. The form of this bound resembles bounds from \citet{blanchard2007statistical} but our result holds for a more general class of kernel. We also state a second bound which introduces a new empirical notion of variance, which shed a new light on the importance of appropriate datasets for efficient KPCA.

\paragraph{Outline.} We introduce our notation and recall existing theoretical results on kernel PCA in \Cref{sec: background}. We state our main results in \Cref{sec: results} and defer the proofs to \Cref{sec: proofs}.

\section{Notation and background}
\label{sec: background}

We let $\mathbb{R}^{m \times n}$ denote the space of matrices $m \times n$ with real entries. The data space is $\mathcal{Z} \subseteq \mathbb{R}^d$. We assume to have access to $S = (z_1, \ldots, z_m) \in \mathcal{Z}^m$, a realisation of the size-$m$ random vector $(Z_1, \ldots, Z_m) \in \mathcal{Z}^m$.
We let $\mathcal{M}_1(\mathcal{Z})$ denote the space of probability distributions over $\mathcal{Z}$ and $\mu \in \mathcal{M}_1(\mathcal{Z})$ stands for the distribution that generates one random example $z \in \mathcal{Z}$. Its empirical counterpart is given by $\hat{\mu} = \frac{1}{m}\sum_{i=1}^{m} \delta_{z_i}$, \emph{i.e.}, the empirical distribution defined by the random sample. We assume the collected sample to be independent and identically distributed (iid): $S \sim \mu^m$, where $\mu^m = \mu\otimes\cdots\otimes\mu$ ($m$ copies).

Denote by $\mathbb{E}_{\nu}[f] = \mathbb{E}_{z \sim\nu}[f(z)] = \int_{\mathcal{Z}} f(z) \nu(dz)$ the expectation under $\nu \in \mathcal{M}_{1}(\mathcal{Z})$, for $f : \mathcal{Z}\to\mathbb{R}$.
We denote by $\mathcal{H}$ a (separable) Hilbert space, equipped with an inner product $\langle \cdot,\cdot \rangle$. We let $\| u \| = \langle u,u \rangle^{1/2}$ be the norm of $u \in \mathcal{H}$.
The operator $P_V : \mathcal{H} \to \mathcal{H}$ is the orthogonal projection onto a subspace $V$, and
$P_{v} = P_{\operatorname{span}\{ v \}}$. We also denote by $P_V^\bot$ the projection on $V^\bot$, the orthogonal of $V$. In what follows, $\mathcal{F}$ is a set of predictors, and $\pi,\pi^0 \in \mathcal{M}_1(\mathcal{F})$ represent probability distributions over $\mathcal{F}$. Finally, $\mathbb{E}_{\pi}[L] = \mathbb{E}_{f \sim\pi}[L(f)] = \int_{\mathcal{F}} L(f) \pi(df)$ is the expectation under $\pi \in \mathcal{M}_1(\mathcal{F})$, for $L : \mathcal{F}\to\mathbb{R}$.

\paragraph{On Reproducing Kernel Hilbert Spaces (RKHS).} We recall results from \citet{hein2004kernels} on the links between RKHS and different mathematical structures.
The key idea is that while data belongs to a data space $\mathcal{Z}$, a kernel function $\kappa : \mathcal{Z}\times\mathcal{Z} \to \mathbb{R}$ implicitly embeds data into a Hilbert space (of real-valued functions), where there is an abundance of structure to exploit. Such a function $\kappa$ is required to be \emph{symmetric} in the sense that $\kappa(z_1,z_2) = \kappa(z_2,z_1)$ for all $z_1,z_2 \in \mathcal{Z}$.

\begin{definition}[PSD kernels]
A symmetric real-valued function $\kappa:\mathcal{Z}\times\mathcal{Z}\rightarrow \mathbb{R}$ is said to be a \emph{positive semi definite} (PSD) kernel if
$\forall n\geq 1$, $\forall z_1,...,z_n \in\mathcal{Z}$, $\forall c_1,...,c_n\in\mathbb{R}$:
\begin{align*}
\sum_{i,j=1}^n c_i c_j \kappa(z_i,z_j)\geq 0.
\end{align*}
If the inequality is strict (for non-zero coefficients $c_1,\ldots,c_n$), then the kernel is said to be \emph{positive definite} (PD).
\end{definition}
For instance, polynomial kernels $\kappa(x,y)= (x^{T}y+r)^n$, and Gaussian kernels  $\kappa(x,y)=\exp(-||x-y||^2/2\sigma^2)$ (for $n\geq 1,(x,y)\in(\mathbb{R}^d)^2, r\geq 0,\sigma>0$) are PD kernels.
We define below a particular class of kernels which will be useful later.

\begin{definition}
\label{def: translation_inv_kernel}
A kernel $\kappa$ is said to be \emph{translation-invariant} if there exists a fuction $g: \mathcal{Z}\rightarrow \mathbb{R}$ such that for all $(x,y)$, $\kappa(x,y)= g(x-y)$.
\end{definition}
Note that in particular, Gaussian kernels are translation-invariant.

\begin{definition}[RKHS]
A \emph{reproducing kernel Hilbert space} (RKHS) on $\mathcal{Z}$ is a Hilbert space $\mathcal{H} \subset \mathbb{R}^\mathcal{Z}$ (functions from $\mathcal{Z}$ to $\mathbb{R}$) where all evaluation functionals $\delta_z : \mathcal{H} \rightarrow \mathbb{R}$, defined by $\delta_z(f) = f(z)$, are continuous.
\end{definition}
Note that, since the evaluation functionals are linear, an equivalent condition to continuity (of all the $\delta_z$'s) is that for every $z\in\mathcal{Z}$, there
exists $M_z < +\infty$ such that
\[
\forall f \in \mathcal{H}, \hspace{3mm} |f(z)|\leq M_z ||f||.
\]
This condition is the so-called \emph{reproducing property}.
When $\mathcal{H}$ is an RKHS over $\mathcal{Z}$, there is a kernel $\kappa:\mathcal{Z}\times\mathcal{Z}\rightarrow \mathbb{R}$ and a mapping $\psi:\mathcal{Z}\to\mathcal{H}$ such that $\kappa(z_1,z_2) = \langle \psi(z_1),\psi(z_2) \rangle_{\mathcal{H}}$. Intuitively: the `feature mapping' $\psi$ maps data points $z \in \mathcal{Z}$ to `feature vectors' $\psi(z) \in \mathcal{H}$, while the kernel computes the inner products between those feature vectors without needing explicit knowledge of $\psi$.

The following key theorem (known as the Moore-Aronszajn theorem) links PD kernels and RKHS.

\begin{theorem}[\citealp{aronszajn1950}]
If $\kappa$ is a positive definite kernel, then there exists a unique reproducing kernel Hilbert space $\mathcal{H}$ whose kernel is $\kappa$.
\end{theorem}
\begin{proof}[Proof (from \citealp{hein2004kernels})]
from a PD kernel $\kappa$, we build an RKHS from the pre-Hilbert space
$$
V=\operatorname{span}\left\{\kappa(z,.)\mid z\in\mathcal{Z} \right\}.
$$
We endow V with the following inner product:
\begin{equation}\label{remark_RKHS_finite_data_space_X}
\left\langle \sum_{i}a_i\kappa(z_i,.), \sum_j b_j \kappa(z_j,.)  \right\rangle_V = \sum_{i,j} a_i b_j \kappa(z_i,z_j) .
\end{equation}
It can be shown that this is indeed a well-defined inner product.
Thus, the rest of the proof consists in the completion of $V$ into an Hilbert space (of functions) verifying the reproducing property, and the verification of the uniqueness of such an Hilbert space.
\end{proof}
An important special case is when $|\mathcal{Z}|< +\infty$, then $V$ is a finite-dimensional vector space. Thus if we endow it with an inner product, $V$ is already an Hilbert space (it already contains every pointwise limits of all Cauchy sequences of elements of $V$). As a consequence, the associated RKHS is finite-dimensional in this case.

\begin{definition}[Aronszajn mapping]
\label{aronszajn_mapping}
For a fixed PD kernel $\kappa$, we define the \emph{Aronszajn mapping}
    $\psi: \mathcal{Z}\rightarrow \left(\mathcal{H},\langle \cdot,\cdot\rangle\right)$ such that
    \[
    \forall z \in\mathcal{Z}, \hspace{3mm} \psi(z)= \kappa(z,.)    ,
    \]
    where we denote by $\mathcal{H}$ the RKHS given by the Moore-Aronszajn theorem and $\langle\cdot,\cdot\rangle$ is the inner product given in  \eqref{remark_RKHS_finite_data_space_X}.
    In the sequel, we refer to the Aronszajn mapping as the pair $(\psi,\mathcal{H})$ when it is important to highlight the space $\mathcal{H}$ into which $\psi$ embeds the data.
    \end{definition}

When it comes to embedding points of $\mathcal{Z}$ into a Hilbert space through a feature mapping $\psi$, several approaches have been considered \citep[see][Section 3.1]{hein2004kernels}. The Aronszajn mapping is one choice among many.

\paragraph{On kernel PCA.} We present here the results from \citet{shawe-taylor2005eigenspectrum}. Fix a PD kernel $\kappa$. We denote by $(\psi,\mathcal{H})$ the Aronszajn mapping of $\mathcal{Z}$ into $\mathcal{H}$.

\begin{definition}
\label{d:gram_matrix}
The kernel Gram matrix of a dataset $S=(z_1,\ldots,z_m) \in \mathcal{Z}^m$ is the element $K(S)$ of $\mathbb{R}^{m \times m}$ defined as
\[
K(S)= (\kappa( z_i,z_j))_{i,j}.
\]
Whenever there is no ambiguity, we shorten this notation to $K = (\kappa( z_i,z_j))_{i,j}$.
\end{definition}

The goal of kernel PCA is to analyse $K$ by putting the intricate data sample of size $m$ from the set $\mathcal{Z}$ into $\mathcal{H}$, where data are properly separated, and then find a small (in terms of dimension) subspace $V$ of $\mathcal{H}$ which catches the major part of the information contained in the data.
We define $\mu \in \mathcal{M}_1(\mathcal{Z})$, a probability measure over $\mathcal{Z}$, as the distribution representing the way data are spread out over $\mathcal{Z}$. In other words, we want to find a subspace $V \subseteq \mathcal{H}$ such as
\[
\forall z\in \mathcal{Z}, \hspace{3mm}
\left| ||P_V(\psi(z)||^2 - ||\psi(z)||^2 \right| \approx 0,
\]
where $P_V$ is the orthogonal projection over the subspace $V$. We use the notation $[m] = \{ 1,\ldots,m \}$.
Recall that $\psi$ and $\mathcal{H}$ are defined such that we can express the elements of $K$ as a scalar product in $\mathcal{H}$:
\[
\forall i,j \in [m], \hspace{3mm}
K_{i,j}= \kappa(z_i,z_j) = \langle \psi(z_i),\psi(z_j)\rangle_\mathcal{H}.
\]

\begin{definition}
For any probability distribution $\nu$ over $\mathcal{Z}$,
we define the self-adjoint operator on $L^2(\mathcal{Z},\nu)$ associated to the kernel function $\kappa$ as:
\[
\mathcal{K}_\nu (f)(z) = \int_{\mathcal{Z}} f(z') \kappa(z,z')\nu(dz').
\]
\end{definition}

\begin{definition}
\label{d:e_values}
We use the following conventions:
\begin{itemize}
    \item If $\mu$ is the data-generating distribution, then we rename $\mathcal{K} := \mathcal{K}_{\mu}$.
    \item If $\hat{\mu}$ is the empirical (uniform) distribution of our $m$-sample $(x_i)_i$, then we name $\hat{\mathcal{K}} := \mathcal{K}_{\hat{\mu}}$.
    \item $\lambda_1\geq \lambda_2\geq\cdots$ are the eigenvalues of the operator $\mathcal{K}$.
    \item $\hat{\lambda}_1\geq\cdots\geq \hat{\lambda}_m\geq 0$ are the eigenvalues of the kernel matrix $K$.
\end{itemize}
\end{definition}

More generally, let $\lambda_1(A) \geq \lambda_2(A) \geq \cdots$ be the eigenvalues of a matrix $A$, or a linear operator $A$. Then in Definition \ref{d:e_values}, we use the shortcuts $\lambda_i = \lambda_i(\mathcal{K})$ and $\hat{\lambda}_i = \lambda_i(K)$.
Notice that for all $i \in\{1,\dots,m\}$, we have $\lambda_i(\hat{\mathcal{K}})= \frac{\hat{\lambda}_i}{m}$.

\begin{definition}
\label{d:sums}
For a given sequence of real scalars $(a_i)_{i \ge 1}$ of length $m$, which may be infinity, we define for any $k$ the initial sum and the tail sum as
\[
a^{\leq k}:= \sum_{i= 1}^k a_i
\hspace{3mm}\text{ and}\hspace{3mm}
a^{>k}:= \sum_{i= k+1}^m a_i.
\]
\end{definition}

\begin{definition}
\label{d:corr_matrix}
The \emph{sample covariance matrix} of a random dataset $S=(X_1,\ldots,X_m)$ is the element $C(S)$ of $\mathbb{R}^{m \times m}$ defined by
\[
C(S)= \frac{1}{m}\sum_{i=1}^m \psi(X_i)\psi(X_i)',
\]
where $\psi(x)'$ denotes the transpose of $\psi(x)$.
Notice that this is the sample covariance in feature space.
Whenever there is no ambiguity, we shorten $C(S)$ to $C$.
\end{definition}

One could object that $C$ may not be finite-dimensional, because $\mathcal{H}$ is not (in general). However, notice that the subspace of $\mathcal{H}$ spanned by  $\psi(x_1),...,\psi(x_m)$ is always finite-dimensional, hence by choosing a basis of this subspace, $C$ becomes effectively a finite-dimensional square matrix (of size no larger than $m \times m$).

\begin{definition}
\label{def: subspaces}
For any probability distribution $\nu$ over $\mathcal{Z}$, we define $\mathcal{C}_\nu : \mathcal{H} \to \mathcal{H}$ as the mapping $\alpha \mapsto \mathcal{C}_\nu(\alpha)$ where:
\[
\mathcal{C}_\nu(\alpha) = \int_{\mathcal{Z}}\langle \psi(z),\alpha\rangle \psi(z)\nu(dz).
\]
If $\nu$ has density $v(z)$, then we write
$\mathcal{C}_v (\alpha) = \int_{\mathcal{Z}}\langle \psi(z),\alpha\rangle \psi(z)v(z)dx$.
Notice that the eigenvalues of $\mathcal{K}_\nu$ and $\mathcal{C}_\nu$ are the same for any $\nu$, the proof of this fact may be found in \citep{shawe-taylor2005eigenspectrum}.
We similarly define the simplified notations $\mathcal{C} := \mathcal{C}_{\mu}$ (when $\mu$ is the population distribution)
and $\hat{\mathcal{C}} = \mathcal{C}_{\hat{\mu}}$ (when $\hat{\mu}$ is the empirical distribution).
We then define for any $k\in\{1,\dots,m\}$
\begin{itemize}
\item $V_k$ the subspace spanned by the $k$-first eigenvectors of $\mathcal{C}$,
\item $\hat{V_k}(S)$ the subspace spanned by the $k$-first eigenvectors of $\hat{\mathcal{C}}$.
\end{itemize}
\end{definition}
Notice that $\hat{\mathcal{C}}$ coincides with the sample covariance matrix $C$, \emph{i.e.} we have
\[
\hat{\mathcal{C}}(\alpha)=  C\alpha,
\hspace{5mm} \forall \alpha \in \mathcal{H}.
\]
So for any $k>0$, $\hat{V_k}(S)$ is the subspace spanned by the first $k$ eigenvectors of the matrix $C$.

\begin{proposition}[Courant-Fischer-Weyl minimax theorem, Theorem C in \citealp{shawe-taylor2005eigenspectrum}]
\label{p:courant-fischer}
If $(u_i)_i$ are the eigenvectors associated to $(\lambda_i(\mathcal{K}_\nu))_i$ and $V_k$ is the space spanned by the $k$ first eigenvectors:
\[
\lambda_k(\mathcal{K}_\nu)
= \mathbb{E}_\nu[||P_{u_k}(\psi(z))||^2]
= \max_{\dim(V)=k} \min_{0\neq v\in V}\mathbb{E}_\nu[||P_{v}(\psi(z))||^2],
\]
\[
\lambda^{\leq k}(\mathcal{K}_\nu) = \max_{\dim(V)=k} \mathbb{E}_\nu[||P_{V}(\psi(z))||^2],
\]
\[
\lambda^{>k}(\mathcal{K}_\nu) = \min_{\dim(V)=k} \mathbb{E}_\nu[||P_{V}^\bot(\psi(z))||^2].
\]
\end{proposition}

Now we denote by $\mathbb{E_{\mu}}$ the expectation under the true data-generating distribution $\mu$ and by $\mathbb{E}_{\hat{\mu}}$ the expectation under the empirical distribution of an $m$-sample $S$.
Combining the last property with the Courant-Fisher theorem (as stated in \citealp[Theorem C]{shawe-taylor2005eigenspectrum}) gives us the following equalities.

\begin{proposition}
\label{p: characterization_eigenvalues}
We have
\begin{align*}
\mathbb{E}_{\hat{\mu}} [ ||P_{\hat{V_k}}(\psi(z))||^2]
&= \frac 1 m \sum_{i=1}^m ||P_{\hat{V _k}}(\psi(z_i))||^2  = \frac 1 m \sum_{i=1}^k \hat{\lambda}_i = \mathbb{E}_{\hat{\mu}} [ ||\psi(x)||^2 ] - \sum_{i=k+1}^m \hat{\lambda}_i  ,\\
\mathbb{E}_{\mu} [ ||P_{V_k}(\psi(z))||^2]
&= \sum_{i=1}^k \lambda_i = \mathbb{E}_{\mu} [ ||\psi(x)||^2 ] - \sum_{i=k+1}^m \lambda_i .
\end{align*}
\end{proposition}

\paragraph{Previous results} We introduce here several well-known results in KPCA literature we will compare ourselves to. We first recall two important results from \citet{shawe-taylor2005eigenspectrum} and one from \citet{blanchard2007statistical} that will be useful for comparison below.
With the notation introduced in Definition \ref{d:sums}, when projecting onto the subspace $\hat{V_k}(S)$ spanned by the first $k$ eigenvectors of $\hat{\mathcal{C}} = \hat{\mathcal{K}}$,
the tail sum $\lambda^{>k} = \sum_{i>k} \lambda_i$ lower-bounds the expected squared residual.

\begin{theorem}[\citealp{shawe-taylor2005eigenspectrum}, Theorem 1]
\label{th : st_2005_residual}
If we perform PCA in the feature space defined by the Aronszjan mapping $(\psi,\mathcal{H})$ of a kernel $\kappa$, then with probability of at least $1-\delta$ over $m$-samples $S$ we have for all $k \in [m]$ if we project new (independent) data $z$ onto the space $\hat{V}_k(S)$,
the expected square residual satisfies:
\begin{equation*}
\lambda^{>k}  \leq \mathbb{E}_{z\sim\mu}[||P_{\hat{V}_k}^\bot(\psi(z))||^2]
\leq \min_{1\leq \ell \leq k} \left[\frac 1 m\hat{\lambda}^{>\ell}(S) + \frac{1 + \sqrt{\ell}} {\sqrt{m}} \sqrt{\frac 2 m \sum_{i=1}^m \kappa(x_i,x_i)^2}\right]
 + R^2 \sqrt{\frac {18} {m} \log\left(\frac {2m} {\delta}\right)} .
\end{equation*}
This holds under the assumption that $\| \psi(z) \| \leq R$, for any $x \in \mathcal{Z}$.
\end{theorem}

\begin{theorem}[\citealp{shawe-taylor2005eigenspectrum}, Theorem 2]
\label{th: st_2005_proj}
If we perform PCA in the feature space defined by the Aronszjan mapping $(\psi,\mathcal{H})$ of a kernel $\kappa$, then with probability of at least $1-\delta$ over random $m$-samples $S$ we have for all $k \in [m]$ if we project new data $z$ onto the space $\hat{V}_k(S)$, the expected square projection satisfies:
\begin{equation*}
\lambda^{\leq k}  \geq \mathbb{E}_{z\sim\mu}[||P_{\hat{V}_k}(\psi(z))||^2]
\geq \max_{1\leq \ell\leq k} \left[\frac 1 m\hat{\lambda}^{\leq \ell}(S) - \frac{1 + \sqrt{\ell}} {\sqrt{m}} \sqrt{\frac 2 m \sum_{i=1}^m \kappa(x_i,x_i)^2}\right]
 - R^2 \sqrt{\frac {19} {m} \log\left(\frac {2(m+1)} {\delta}\right)} .
\end{equation*}
This holds under the assumption that $\| \psi(z) \| \leq R$, for any $x \in \mathcal{Z}$.
\end{theorem}

Notice that the purpose of those two theorems is to control, by upper and lower bounds, the theoretical averaged squared norm projections of a new data point $x$ into the empirical subspaces $\hat{V}_k$ and $\hat{V}_k^\bot$:  $ \mathbb{E}_{\mu}[||P_{\hat{V}_k}(\psi(z))||^2] $ and $\mathbb{E}_{\mu}[||P_{\hat{V}_k}^\bot(\psi(z))||^2]$. Let us note that for each theorem, only one side of the inequality is empirical (while the other one consists in an unknown quantity, $\lambda
^{\leq k}$ or $\lambda^{>k}$, respectively).
We now state a result from \citet{blanchard2007statistical} which propose a possibly faster convergence rate than the previous ones.

\begin{theorem}[\citealp{blanchard2007statistical}, Theorem 3.4]
\label{th: blanchard_2007}
Assume our kernel $\kappa$ is such that for all $z\in\mathcal{Z}, \kappa(z,z)=R^2$. Then for all $k, m \geq 2, \delta$, with probability at least $1-4\delta$ the following holds:

\begin{multline*}
\left|\mathbb{E}_{z\sim\mu}[||P_{\hat{V}_k}(\psi(z))||^2]-\frac 1 m\hat{\lambda}^{>k}(S)\right| \\ \leq c\left(\sqrt{\frac 1 m \hat{\lambda}^{>k}(S)\left(\rho_{m}(R^2, k, m)+R^2\frac{\log\left( \frac{m}{\delta}  \right)}{m}\right)}+\rho_{m}(R^2,k, m)+R^2\frac{(\log\left( \frac{m}{\delta}  \right) )}{m}\right),
\end{multline*}
where $c$ is a universal constant $\left(c \leq 1.2 \times 10^{5}\right)$. The quantity $\rho_{m}(R^2,k, m)$ varying (for a fixed $d$) between $1/\sqrt{m}$ and $1/m$.
\end{theorem}

Note that this bound holds especially for translation-invariant kernels but may be restrictive otherwise. This bound ensure that if $\frac 1 m\hat{\lambda}^{>k}(S)$ is small and that $\rho_{m}(R^2,k, m)$ behaves as $1/m$, then one has a fast convergence rate of $\log(m)/m$ for the reconstruction error.

\section{Main results}
\label{sec: results}

\subsection{High dimensional projection spaces lead to fast convergence rates.  }

\label{sec: main_result_1}

We start with a bound derived from the PAC-Bayesian theorem of \citet{tolstikhin2013pac}. We recover a bound (\Cref{th: main_result_1}) with a similar form than the one from \citet[][Theorem 3.4]{blanchard2007statistical}, which ensures a fully empirical bound for the case of translation invariant kernels. We claim that our results are strictly more general than \citep[Theorem 3.4]{blanchard2007statistical} as it is valid for any type of kernels. Furthermore, we recover asymptotically the best convergence rate attained in \citep[Theorem 3.4]{blanchard2007statistical}. The originality of our results lies in the fact that our results does not depend directly on the projection dimension $k$.

\begin{theorem}
\label{th: main_result_1}
If we perform PCA in the feature space defined by the Aronszjan mapping $(\psi,\mathcal{H})$ of a kernel $\kappa$, then, for a fixed $k\in [m]$, with probability of at least $1-\delta$ over an $m$-sample $S$, if we project new data $z$ onto the space $\hat{V}_k(S)$, we have

\begin{equation*}
\mathbb{E}_\mu[|| P_{\hat{V}_k(S)}^\bot(\psi(x)||^2]  \leq  \frac{1}{m}\sum_{i=k+1}^m \hat{\lambda}_i +R^2\sqrt{\frac{ \frac{2}{m}\sum_{i=k+1}^m \hat{\lambda}_i\cdot\varepsilon_{m,\delta}}{m}}  + 2R^2\frac{\varepsilon_{m,\delta}}{m}
\end{equation*}
and
\begin{equation*}
\mathbb{E}_\mu[ ||P_{\hat{V}_k(S)}(\psi(x)||^2]  \geq  \frac{1}{m}\sum_{i=1}^k \hat{\lambda}_i - R^2\sqrt{\frac{ \frac{2}{m}\sum_{i=k+1}^m \hat{\lambda}_i\cdot\varepsilon_{m,\delta}}{m}} -2R^2\frac{\varepsilon_{m,\delta}}{m} - \sqrt{\frac{\mathbb{V}_m(S)\log \frac{4}{\delta}}{m}} - R^2\frac{7\log(\frac{4}{\delta})}{3(m-1)},
\end{equation*}
where $R:= \max_{z\in\mathcal{Z}} ||\psi(z)||$, $\varepsilon_{m,\delta} = \log(\alpha(m))+\log \frac{4 \sqrt{m}}{\delta}$ with $\alpha(m)=\binom{|\mathcal{Z}|+m-1}{|\mathcal{Z}|-1}$ and
\begin{equation*}
\mathbb{V}_m(S) = \frac{1}{m(m-1)} \sum_{1\leq i<j\leq m } (\kappa(x_i,x_i)-\kappa(x_j,x_j))^2 .
\end{equation*}
\end{theorem}

\begin{remark}
Note the asymmetry in our results: we have a better control on the information that is not retained by kernel PCA than on the one caught by this method. This asymetry comes from the fact we did not restrict ourselves to translation invariant kernels (or the slight weaker assumption of \citealp[Theorem 3.4]{blanchard2007statistical}).
\end{remark}

\paragraph{Comparison with \citep{shawe-taylor2005eigenspectrum}}
There are in our bound (as in \citealp{shawe-taylor2005eigenspectrum}) two different kinds of terms: those quantifying the quality of our subspace obtained through the kernel PCA algorithm and those which relate to the interaction of the chosen kernel with its associated RKHS.
\begin{itemize}
    \item  First, the term $ \frac{2}{m}\sum_{i=k+1}^m \hat{\lambda}_i$ explains how accurately the information is caught by kernel PCA. \citet{shawe-taylor2005eigenspectrum} proposed a similar term only involving the dimension of projection $k$. Thus our term appears to be tighter than theirs because it shows explicitely the efficiency gain of kernel PCA while reaching higher-dimensional projection subspaces.
    \item Second, our variance term $\mathbb{V}_m(S)$ can be compared to $\frac{2}{m} \sum_{i=1}^m \kappa(x_i,x_i)^2$ in \citet{shawe-taylor2005eigenspectrum}. Both these terms describe how the kernel is interacting with its RKHS. The term exhibited by \citet{shawe-taylor2005eigenspectrum} is the trace of a particular Hilbertian operator which is assessing the impact of the data repartition within the RKHS (recall that for $x\in\mathcal{X}, \|\psi(x)\|^2=\kappa(x,x)$): the more the data is far from the origin, the more costly this term is so this seems to favour kernels which gather data through the origin. On the contrary, our variance term indicates that kernels which associate a similar norm to most of the points are relevant for kernel PCA, which is striclty more general because it allow to have points diametrically opposed on a same sphere.
\end{itemize}

\paragraph{Comparison with \citet[Theorem 3.4]{blanchard2007statistical}}
First, let us notice that \citep[Theorem 3.4]{blanchard2007statistical} only focuses on the control of $  \mathbb{E}_\mu[|| P_{\hat{V}_k(S)}^\bot(\psi(x)||^2]$ while we also give a control of $\mathbb{E}_\mu[|| P_{\hat{V}_k(S)}^\bot(\psi(x)||^2]$.
So our comparison will not take into account our second bound with the variance term which is beyond the scope of \citet{blanchard2007statistical}. Authors acknowledge having a convergence rate varying between
$\frac{\sqrt{\frac{2}{m}\sum_{i=k+1}^m \hat{\lambda}_i}}{m^{1/4}} \quad \text{and} \quad\sqrt{\frac{\frac{2}{m}\sum_{i=k+1}^m \hat{\lambda}_i \log(m)}{m}},$
whatever the size of the sample, while we have a fixed rate of
$\sqrt{\frac{\frac{2}{m}\sum_{i=k+1}^m \hat{\lambda}_i\log(m\alpha(m))}{m}}.$
However, thanks to \Cref{prop: asymptotic_behavior_alpha}, we asymptotically recover the best convergence rate of
$\sqrt{\frac{\frac{2}{m}\sum_{i=k+1}^m \hat{\lambda}_i \log(m)}{m}}$
without involving a potentially big universal constant $c$ (\Cref{th: blanchard_2007}).
We follow the trace of \citet[Theorem 3.4]{blanchard2007statistical} by keeping the partial sum of eigenvalues under a square root. We are even more precise in a way: contrary to \citet{blanchard2007statistical}, our bound does not depend explicitely on the projection dimension $k$ (indeed, the term $\rho_m(R^2,k,m)$ in \Cref{th: blanchard_2007} depends explicitely on $k$).
This dimension only appears in the quantity $ \frac{2}{m}\sum_{i=k+1}^m \hat{\lambda}_i$. Thus the projection dimension is not playing a crucial role: its influence can be fully evaluated through the output of the KPCA algorithm.

Last but not least let us stress that in \citet{blanchard2007statistical}, a fully empirical bound for localised fast rates is reached only for translation invariant kernels (as defined in \Cref{def: translation_inv_kernel}) while our bound holds for any type of kernels.

For the sake of completeness, we show below a corollary of our main result which gives an explicit condition on the residual sum of empirical eigenvalues to attain a fast convergence rate.

\begin{corollary}
\label{cor: invariant_kernels_1}
Under the conditions stated in \Cref{th: main_result_1}, if a kernel is translation invariant and if $k$ is big enough such that $\frac{2}{m}\sum_{i=k+1}^m \hat{\lambda}_i \approx \frac{\log(\alpha(m)}{m}$ then one has with probability at least $1-\delta$:

\begin{equation*}
\mathbb{E}_\mu[|| P_{\hat{V}_k(S)}^\bot(\psi(x)||^2]  \leq  \frac{1}{m}\sum_{i=k+1}^m \hat{\lambda}_i +O\left(\frac{\log(\alpha(m)}{m}\right)
\end{equation*}
and
\begin{equation*}
\mathbb{E}_\mu[ ||P_{\hat{V}_k(S)}(\psi(x)||^2]  \geq  \frac{1}{m}\sum_{i=1}^k \hat{\lambda}_i - O\left(\frac{\log(\alpha(m)}{m}\right).
\end{equation*}
Asymptotically, $\frac{\log(\alpha(m))}{m} \sim |\mathcal{Z}|\frac{\log(m)}{m}$.
\end{corollary}

\subsection{Meaningful data for kernel PCA lead to faster convergence rates}
\label{sec: main_result_2}

We now exploit \citet[Corollary 1]{mhammedi19pac} to obtain a new type of variance term for kernel PCA. We first split our $m$-sample $S= (z_1,...,z_m)$ onto two samples $S_1,S_2$ of same size\footnote{Assume that $m$ is even without loss of generality.} $m/2$ ($S_1$ contains the $m/2$ first points of $S$, $S_2$ holds the rest).

We fix for $k,l\in [m]^2$
\begin{equation*}
R_{m,\ell}^{\prime} := \frac{1}{m} \sum_{i=1}^{m/2} \left(||\psi(z_i)||^2 - ||P_{\hat{V}_\ell(S_2)}(\psi(z_i)||^2\right)^{2} + \frac{1}{m} \sum_{j=m/2+1}^{m} \left(||\psi(z_j)||^2 - ||P_{\hat{V}_\ell(S_1)}(\psi(z_j)||^2
\right)^{2},
\end{equation*}
and
\begin{equation*}
R_{m,k,\ell}  :=\frac{1}{m} \sum_{i=1}^{m/2}\left(||P_{\hat{V}_k(S)}(\psi(z_i)||^2 - ||P_{\hat{V}_\ell(S_2)}(\psi(z_i)||^2\right)^{2} +\frac{1}{m}\sum_{j=m/2+1}^{m} \left(||P_{\hat{V}_k(S)}(\psi(z_j)||^2 - ||P_{\hat{V}_\ell(S_1)}(\psi(z_j)||^2\right)^{2} .
\end{equation*}

\begin{theorem}
\label{th: main_result_2}
If we perform PCA in the feature space defined by the Aronszjan mapping $(\psi,\mathcal{H})$ of a kernel $\kappa$, then, for a fixed $k\in [m]$, there exists $\mathcal{P}, \mathcal{A}, \mathcal{C}>0$, such that with probability of at least $1-\delta$ over an $m$-sample $S= (z_1,...,z_m)$, if we project new  data $z$ onto the space $\hat{V}_k(S)$, we have

\begin{equation*}
\mathbb{E}_{z\sim \mu}[||P_{\hat{V}_k^\bot(S)}(\psi(z))||^2]  \leq \frac{1}{m}\sum_{i=k+1}^m \hat{\lambda}_i + \min_{\ell\in[m]} \left[ \mathcal{P} \cdot \sqrt{\frac{R_{m,k,\ell} \cdot\varepsilon_{\delta, m}}{m}} +\mathcal{A} \cdot \frac{\varepsilon_{\delta, m}}{m}
+\mathcal{C} \cdot \sqrt{\frac{R_{m,\ell}^{\prime} \cdot \varepsilon_{\delta, m}}{m}} \right],
\end{equation*}
and
\begin{multline*}
\mathbb{E}_{z\sim \mu}[||P_{\hat{V}_k(S)}(\psi(z)||^2] \\ \geq \frac{1}{m}\sum_{i=1}^k \hat{\lambda}_i - \min_{\ell \in [m]} \left[ \mathcal{P} \cdot \sqrt{\frac{R_{m,k,\ell} \cdot\varepsilon_{\delta, m}}{m}} +\mathcal{A} \cdot \frac{\varepsilon_{\delta, m}}{m}
+\mathcal{C} \cdot \sqrt{\frac{R_{m,\ell}^{\prime} \cdot \varepsilon_{\delta, m}}{m}} \right]
- \sqrt{\frac{\mathbb{V}_m(S)\log \frac{2}{\delta}}{m}} - \frac{7\log(\frac{2}{\delta})}{3(m-1)},
\end{multline*}
where $\varepsilon_{m,\delta}= O\left( \log(\alpha(m)) + \log\left( \frac{\log(m)}{\delta}   \right)\right)$, $\alpha(m)=\binom{|\mathcal{Z}|+m-1}{|\mathcal{Z}|-1}$ and
\begin{align*}
\mathbb{V}_m(S) & = \frac{1}{m(m-1)} \sum_{1\leq i<j\leq m } (\kappa(z_i,z_i)-\kappa(z_j,z_j))^2.
\end{align*}
\end{theorem}
First of all, notice that as in \Cref{sec: main_result_1}, the projection dimension $k$ only intervenes in our bound through the projection on some $\hat{V}_k(S')$ with $S'$ being $S$, $S_1$ or $S_2$. This is a significant improvement that ensures empirical guarantees directly from the output of KPCA. Indeed, in \citet{shawe-taylor2005eigenspectrum}\footnote{See our \Cref{th: st_2005_proj}.}, a minimisation is made using a function which takes directly as argument the dimension $k$, which is not directly linked to the quality of the projections. That is why our approach suggests a tradeoff between $R'_{m,\ell}$ and $R_{m,k,\ell}$ which directly involves informative quantities about the quality of data. We now describe what we call \emph{quality} and \emph{coherence} (capturing the notion of \emph{meaningfulness}) of a dataset.
\begin{itemize}
  \item \emph{Quality.} $R'_{m,\ell}$ represents how qualitative are our data. More precisely, can we still have meaningful projections on new data if we were only using half of the dataset? Indeed, $R'_{m,\ell}$ assesses, for each half-sample $S_i$ of the data, the quality of the projection of the other half onto the subspace of dimension $\ell$ generated by $S_i$. Thus $R'_{m,\ell}$ is hinting at how good can kernel PCA ve if we only use half of data.
  \item \emph{Coherence.} $R_{m,k,\ell}$ is a new variance term which evaluates the coherence of our dataset in the sense that if we choose to remove half of data, are the projections geometrically coherent (in the sense of the squared norm) with those obtained with the entire dataset?
\end{itemize}
We also denote that for each $k$, our bound contains a minimum over $\ell$ but the minimised function still varies with $k$. A good tradeoff is not necessarily attained for $\ell=k$ because we lose information by throwing away half of the information. Hence the need to consider all the available $(\hat{V}_\ell(S_i))_{i=1,2}$ to check the meaningfulness of our data.

Finally notice that if data is meaningful in the sense described above, we can attain a fast convergence rate.

\begin{corollary}
\label{cor: fast_rate_2}
Perform KPCA in the feature space defined by the Aronszjan mapping $(\psi,\mathcal{H})$ of a kernel $\kappa$. Let $k\in [m]$, assume data is meaningful, \emph{i.e.} there exists $\ell\in[m]$ such that $R_{m,k,l} = O(\frac{\varepsilon_{\delta, m}}{m})$ (coherence), $R'_{m,\ell}= O(\frac{\varepsilon_{\delta, m}}{m})$ (quality), then there exists $\mathcal{P}$, $\mathcal{A}$, $\mathcal{C}>0$,
such that with probability at least $1-\delta$ over an $m$-sample $S= (z_1,...,z_m)$, by projecting new data $z$ onto the space $\hat{V}_k(S)$, we have
\begin{equation*}
\mathbb{E}_{z\sim \mu}[||P_{\hat{V}_k^\bot(S)}(\psi(z))||^2]  \leq \frac{1}{m}\sum_{i=k+1}^m \hat{\lambda}_i + O(\frac{\varepsilon_{\delta, m}}{m}),
\end{equation*}
and if the kernel is translation invariant, we have
\begin{equation*}
\mathbb{E}_{z\sim \mu}[||P_{\hat{V}_k(S)}(\psi(z)||^2]  \geq \frac{1}{m}\sum_{i=1}^k \hat{\lambda}_i - O(\frac{\varepsilon_{\delta, m}}{m}),
\end{equation*}
with $\varepsilon_{m,\delta}= O\left( \log(\alpha(m)) + \log\left( \frac{\log(m)}{\delta}  \right)\right)$, $\alpha(m)=\binom{|\mathcal{Z}|+m-1}{|\mathcal{Z}|-1}$ and asymptotically, $\frac{\log(\alpha(m)}{m} \sim |\mathcal{Z}|\frac{\log(m)}{m})$.

\end{corollary}

\section{Proofs}
\label{sec: proofs}

\subsection{Elements from the PAC-Bayes theory}

We consider a fixed learning problem with data space $\mathcal{Z}$, set of predictors $\mathcal{F}$, and loss function $\ell : \mathcal{F}\times \mathcal{Z} \rightarrow \mathbb{R}^{+} $.
We denote by $\mu$ a probability distribution over $\mathcal{Z}$,
$S = (z_1,\ldots,z_m)$ is a size-$m$ sample.
We denote as $\Sigma_{\mathcal{F}}$ the considered $\sigma$-algebra on $\mathcal{H}$.
Finally,  we define for any $f\in\mathcal{F}$, $L(f)= \mathbb{E}_{z\sim \mu}[\ell(f,z)]$ and $\hat{L}_m(f)=\frac{1}{m}\sum_{i=1}^m \ell(f,z_i)$.

We define the Gibbs regression rule $G_{Q}$ associated with a distribution $Q$ over $\mathcal{F}$ in the following way: for each point $X$, Gibbs regression rule draws an hypothesis $f$ according to $Q$ and applies it to $X$. The expected loss of Gibbs regression rule is denoted by $L\left(G_{Q}\right)=\mathbb{E}_{f \sim Q}[L(f)]$ and the empirical loss is denoted by $L_{m}\left(G_{Q}\right)=\mathbb{E}_{f \sim Q}\left[L_{m}(f)\right] .$
We use $\mathrm{KL}(Q \| P)=\mathbb{E}_{f \sim Q}\left[\ln \frac{Q(f)}{P(f)}\right]$ to denote the Kullback-Leibler divergence between two probability distributions $P$ and $Q$.
$P$ denotes a prior distribution over $\mathcal{F}$ in the sense that $P$ does not depend of our dataset $S$.
Let us highlight that in supervised learning, for instance, $\mathcal{Z} = \mathcal{X}\times\mathcal{Y}$
where $\mathcal{X} \subset \mathbb{R}^d$ is a space of inputs, and $\mathcal{Y}$ a space of labels. In this case predictors are functions $f:\mathcal{X}\to\mathcal{Y}$. One may be interested in applying PCA over the input space $\mathcal{X}$ to reduce the input dimension.

We now can provide our two theorem of interest. The first one is extracted from \citet{tolstikhin2013pac}, the second one comes from \citet{mhammedi19pac}.

\begin{theorem}[\citealp{tolstikhin2013pac}, Sec 2.2]
\label{th: TS_pac_bayes}
For any loss function in $[0;1]$, any prior $P$, with probability $1-\delta$ over the sample $S$, it holds for any posterior $Q$:

\begin{equation*}
L\left(G_{Q}\right)  \leq L_{m}\left(G_{Q}\right)+\sqrt{\frac{2 L_{m}\left(G_{Q}\right)\left(\mathrm{KL}(Q \| P)+\log \frac{2 \sqrt{m}}{\delta}\right)}{m}}
+\frac{2\left(\mathrm{KL}(Q \| P)+\log \frac{2 \sqrt{m}}{\delta}\right)}{m} .
\end{equation*}
\end{theorem}

Before stating the bound from \citet{mhammedi19pac}, we define for a $m$-sized dataset $S$ and a fixed integer $1\leq n \leq m$, the quantites $S_{\leq n} := (z_1,...,z_n)$ and $S_{>n}:= (z_{n+1},...,z_m)$. We also define $\hat{f}(S')$, for any sample $S'$, to be a measurable function in $S'$.
More generally, $\hat{f}: \bigcup_{i=1}^{m} \mathcal{Z}^{i} \rightarrow \mathcal{F}$ is a deterministic estimator. For instance $\hat{f}$ can be obtained through stochastic gradient descent (SGD) and $\hat{f}(S)$ is the predictor obtained after $|S|$ iterations of SGD on the sample $S$.

\begin{theorem}[\citealp{mhammedi19pac}, Cor. 1]
\label{th: mhameddi_pac_bayes}
Let $S$ be a $m$-sized sample, and $1 \leq n<m$. Let  $P(S_{\leq n}), P(S_{>n})$ be two priors depending respectively on $S_{\leq n}, S_{>n}$.
For any deterministic estimator $\hat{f}: \bigcup_{i=1}^{m} \mathcal{Z}^{i} \rightarrow \mathcal{H}$ ,for any bounded loss function, there exists $\mathcal{P}, \mathcal{A}, \mathcal{C}>0$, such that with probability at least $1-\delta$ over the sample $S=(z_{1}, \ldots, z_{m})$, for any posterior $Q$ it holds that
\begin{equation*}
L\left(G_{P_{m}}\right)-L_{m}\left(G_{P_{m}}\right) \leq \mathcal{P} \cdot \sqrt{\frac{R_{m} \cdot\left(\operatorname{COMP}_{m}+\varepsilon_{\delta, m}\right)}{m}}+\mathcal{A} \cdot \frac{\operatorname{COMP}_{m}+\varepsilon_{\delta, m}}{m}
 +\mathcal{C} \cdot \sqrt{\frac{R_{m}^{\prime} \cdot \varepsilon_{\delta, m}}{m}},
\end{equation*}
where
\begin{equation*}
\varepsilon_{\delta,m} \propto \log \left( \frac{\log(m)}{\delta}   \right),
\end{equation*}
\begin{equation*}
\operatorname{COMP}_{m}=\operatorname{KL}\left(Q \| P\left(S_{\leq n}\right)\right)+\operatorname{KL}\left(Q \| P\left(S_{>n}\right)\right),
\end{equation*}
\begin{equation*}
R_{m}^{\prime}:=\frac{1}{m} \sum_{i=1}^{n} \ell_{\hat{f}\left(S_{>n}\right)}\left(S_{i}\right)^{2}+\frac{1}{m} \sum_{j=n+1}^{m} \ell_{\hat{f}\left(S_{\leq n}\right)}\left(S_{j}\right)^{2},
\end{equation*}
\begin{equation*}
R_{m}:=\frac{1}{m} \mathbb{E}_{f \sim P_{m}}\left[\sum_{i=1}^{n}\left(\ell_{f}\left(S_{i}\right)-\ell_{\hat{f}\left(S_{>n}\right)}\left(S_{i}\right)\right)^{2}  +\sum_{j=n+1}^{m}\left(\ell_{f}\left(S_{j}\right)-\ell_{\hat{f}\left(S_{\leq n}\right)}\left(S_{j}\right)\right)^{2}\right] .
\end{equation*}

\end{theorem}

\subsection{Learning theory framework for kernel PCA}
\label{sec: LT_framework_kpca}

From now on, we work under the assumption $|\mathcal{Z}|< \infty$. In this case, $\mathcal{H}$ is finite-dimensional of dimension $N_\mathcal{H}= |\mathcal{Z}|$.

We need a tuple $(\mathcal{Z},\mathcal{F},\ell)$ where $\mathcal{Z}$ is a data space, $\mathcal{F}$ a predictor space, and a $\ell$ loss function. For now we only have a (finite) data space $\mathcal{Z}$.

From \Cref{sec: background}, we know that kernel PCA generates meaningful data-dependent subspaces of our RKHS $\mathcal{H}$ ($\hat{V}_k(S))_{k=1..m}$, from a sample $S$ and its correlation matrix $C(S)$. Mathematically, one defines a mapping $C(S)\rightarrow (\hat{V}_1(S)...,\hat{V}_m(S)$ which associates $S$ to the $m$-tuple of subspaces spanned by the eigenvectors of $C(S)$.

\begin{remark}
Remark that we state $C(S)$ is a $N_\mathcal{H}\times N_\mathcal{H}$ matrix but can be assimilated to a particular $m\times m$ matrix. Actually, we know that $r=rank(C(S))\leq m$. If this $r$ is effectively equal to $m$ then all the eigenvectors are properly defined. Otherwise, we only have $r$-properly defined subspaces. We then use the axiom of choice to choose a $m-r$ linearly independent family of $\hat{V}_r(S)^\bot$ to complete properly this $r$-tuple onto a $m$-tuple.
\end{remark}

From a learning theory perspective, those subspaces are predictors, hence, if we fix the dimension $k$ of our projected subspace, we have the following predictor space:
\[ \mathcal{F}_k := \left\{ \hat{V_k}(S) \mid S\in \mathcal{Z}^m  \right\}.  \]
We also introduce the space of subspaces tuples which will be useful:
\[ \mathcal{F} := \left\{ (\hat{V_1}(S),...,\hat{V}_m(S) \mid S\in \mathcal{Z}^m  \right\}.  \]
We then consider the following loss function for any subspace $V$ of $\mathcal{H}$ and any data $z\in\mathcal{Z}$, we have
\[ \ell(V,x):= ||\psi(z)||^2 - ||P_V(\psi(z)||^2 = \| P_V^\bot(\psi(z))\|^2.   \]

Before applying the PAC-Bayesian bound, we focus on the cardinality of $\mathcal{F}_k$ for any $k\geq0$. This will be helpful to bound the Kullback-Leibler (KL) divergence term.

\subsection{Combinatorics preliminaries}

\label{sec: combinatorics}

We start our study with the preliminary remark that for any $k, |\mathcal{F}_k| \leq |\mathcal{F}|$.  Indeed for a fixed $k$ ad any element $V\in \mathcal{F}_k$, there exists a $m$-sized sample $S$ such that $V= \hat{V}_k(S)$. We then consider the application $ \hat{V}_k(S) \rightarrow (V_1(S)...,V_m(S))$ which is an injective function.
We then propose the following lemma.
\begin{lemma}
\label{l: combinatorics_KL}
For any $m\geq 0$, any $1\leq k\leq m$ the cardinal of $\mathcal{F}_k$ is bounded as follows:

\[ |\mathcal{F}_k | \leq\binom{|\mathcal{Z}|+m-1}{|\mathcal{Z}|-1}:= \alpha(m).    \]
\end{lemma}

\begin{proof}
Let $m\geq 0$ and $1\leq k \leq m$.  As precised earlier in this subsection, $|\mathcal{F}_k| \leq |\mathcal{F}|$. Our goal is now to bound the number of $m$-tuples in $\mathcal{F}$.
A key remark is that several datasets are leading to the same covariance matrix. For instance if $S_1=\{\psi(x_1),\psi(x_2),\psi(x_1)\},\; S_2 =  \{\psi(x_1),\psi(x_1),\psi(x_2)\} $, then $C(S_1)=C(S_2)$.

What actually matters is the $|\mathcal{Z}|$-tuple $n(S)=(n_1,...,n_{|\mathcal{Z}|})$ of nonnegative integers which informs us how many time each element of $\mathcal{Z}$ will appear in $S$ (n($S$) can be seen as the histogram of the number of occurences of every $z\in\mathcal{Z}$ in $S$) . Thus, a necessary condition for two samples $S_1,S_2$ to have different covariance matrices is $n(S_1)\neq n(S_2)$. This leads us to the bound

\[  |\mathcal{F}|\leq \left | \left\{ (n_1,...,n_{|\mathcal{Z}|}) \mid \forall i, n_i\geq 0,\; \sum_{i=1}^{|\mathcal{Z}|} n_i=m \right\} \right|.       \]

One recognises the last quantity as the number of possibilities to put $m$ balls in $|\mathcal{Z}|$ boxes. This cardinal is obtained through a "balls into boxes" method (see \citealp[p. 26]{stanley2011enumerative} for more details). We know that
\[  |\mathcal{F}| \leq \binom{|\mathcal{Z}|+m-1}{|\mathcal{Z}|-1}.       \]
Using that $|\mathcal{F}_k| \leq |\mathcal{F}|$ concludes the proof.
\end{proof}
The following proposition shows that this bound on $|\mathcal{F}_k|$ is asymptotically strong.
\begin{proposition}
\label{prop: asymptotic_behavior_alpha}
On can control the behavior of  $\log(\alpha)$ as $m$ goes to infinity:
\[ \log(\alpha(m))\sim (|\mathcal{Z}|-1)\log(m). \]
\end{proposition}

\begin{proof}
We have
\begin{align*}
\log(\alpha(m)) & = \log\left( \binom{|\mathcal{Z}|+m-1}{m} \right)  \\
& =  \log\left(\frac{ \Pi_{i=0}^{m-1} |\mathcal{Z}| +i} {m!}  \right)\\
& = \sum_{i=1}^{m} \log\left(1 + \frac{|\mathcal{Z}|-1}{i}\right).
\end{align*}
Furthermore, when $i$ goes to infinity, $\log\left(1 + \frac{|\mathcal{Z}|-1}{i}\right)\sim \frac{|\mathcal{Z}|-1}{i}$. And the series $\sum_{i=1}^m \frac{|\mathcal{Z}|-1}{i}$ is divergent with its general term nonnegative. We finally have, as $m$ goes to infinity:
\[ \sum_{i=1}^{m} \log\left(1 + \frac{|\mathcal{Z}|-1}{i}\right) \operatorname{\sim} \sum_{i=1}^m \frac{|\mathcal{Z}|-1}{i} \sim (|\mathcal{Z}|-1)\log(m).    \]
This concludes the proof.
\end{proof}
We are now able to state and prove our main theorems.

\subsection{Proof of \Cref{th: main_result_1}}

\label{sec: tolstikhin for kpca}

We exploit the framework stated in \Cref{sec: LT_framework_kpca} to obtain the following theorem.

\begin{theorem}
\label{th: main_bounds_TS}
Perform KPCA in the feature space defined by the Aronszjan mapping $(\psi,\mathcal{H})$ of a kernel $\kappa$, then, for a fixed $k\in [m]$, with probability at least $1-\delta$ over an $m$-sample $S$, projecting new data $z$ onto the space $\hat{V}_k(S)$, we have
\begin{equation*}
\mathbb{E}_\mu[|| P_{\hat{V}_k(S)}^\bot(\psi(x)||^2]  \leq \frac{1}{m}\sum_{i=k+1}^m \hat{\lambda}_i +R^2\sqrt{\frac{ \frac{2}{m}\sum_{i=k+1}^m \hat{\lambda}_i\left(\varepsilon_{m,\delta}\right)}{m}}  + R^2\frac{2\left(\varepsilon_{m,\delta}\right)}{m},
\end{equation*}
and
\begin{equation*}
\mathbb{E}_\mu[ ||P_{\hat{V}_k(S)}(\psi(x)||^2]  \geq \frac{1}{m}\sum_{i=1}^k \hat{\lambda}_i - R^2\sqrt{\frac{ \frac{2}{m}\sum_{i=k+1}^m \hat{\lambda}_i\left(\varepsilon_{m,\delta}\right)}{m}} -2R^2\frac{\varepsilon_{m,\delta}}{m} - \sqrt{\frac{\mathbb{V}_m(S)\log \frac{4}{\delta}}{m}} - R^2\frac{7\log(\frac{4}{\delta})}{3(m-1)},
\end{equation*}
where $R:= \max_{z\in\mathcal{Z}} ||\psi(z)||$, $\varepsilon_{m,\delta} = \log(\alpha(m))+\log \frac{4 \sqrt{m}}{\delta}$ with $\alpha(m)=\binom{|\mathcal{Z}|+m-1}{|\mathcal{Z}|-1}$. $\mathbb{V}_m(S)$ is the following variance term:
\begin{equation*}
\mathbb{V}_m(S) = \frac{1}{m(m-1)} \sum_{1\leq i<j\leq m } (\kappa(x_i,x_i)-\kappa(x_j,x_j))^2.
\end{equation*}

\end{theorem}

\begin{proof}
We first fix $k\in\{1..m\}$ and our $m$-sample $S$.
We consider the framework introduced in \Cref{sec: LT_framework_kpca} (i.e the tuple $(\mathcal{Z},\mathcal{F}_k,\ell)$). Our considered prior $P$ is the uniform distribution over $\mathcal{F}_k$
and we take as posterior the Dirac distribution on $\hat{V}_k(S)$. This ensure several points: First $L(G_Q)= \mathbb{E}_{z\sim\mu}[||P_{\hat{V}_k(S)}^\bot(\psi(z))||^2]$ and $L_m(G_Q)= \frac{1}{m}\sum_{i=1}^m||P_{\hat{V}_k}^\bot(\psi(z_i))||^2$.
Second, thanks to \Cref{p: characterization_eigenvalues}, one knows that $\frac{1}{m}\sum_{i=1}^m||P_{\hat{V}_k}^\bot(\psi(z_i))||^2 = \sum_{i=k+1}^m \hat{\lambda}_i$.

We now apply \Cref{th: TS_pac_bayes} to the loss function $\ell/R^2 \in [0;1]$ and we multiply both sides by $R^2$ we then obtain that with probability at least $1-\delta$,
\begin{equation*}
\mathbb{E}_\mu[|| P_{\hat{V}_k(S)}^\bot(\psi(z)||^2]  \leq \frac{1}{m}\sum_{i=k+1}^m \hat{\lambda}_i +R^2\sqrt{\frac{ \frac{2}{m}\sum_{i=k+1}^m \hat{\lambda}_i\left(\mathrm{KL}(Q \| P)+\log \frac{2 \sqrt{m}}{\delta}\right)}{m}}
 +2R^2\frac{\mathrm{KL}(Q \| P)+\log \frac{2 \sqrt{m}}{\delta}}{m}.
\end{equation*}

We now deal with the KL term. To do so, we remark that since $P$ is an uniform distribution on a space of cardinal $|\mathcal{F}_k|$ and $Q$ a Dirac, we have $\operatorname{KL}(Q\|P)= \log\left(|\mathcal{F}_k|\right)$. Thanks to Lemma \ref{l: combinatorics_KL}, we conclude $\KL(Q\|P)\leq \log(\alpha(m))$. This provides the first bound for a fixed $k$.

To obtain the second bound (still for a fixed $k$), we use the fact that for any $(V,z), \|P_V(\psi(z)\|^2 + \|P_V^\bot (\psi(z))\|^2 = \|\psi(z)\|^2 $ and Lemma \ref{l: combinatorics_KL}, we have with probability at least $1-\delta$

\begin{align*}
\mathbb{E}_{z\sim\mu}[ ||P_{\hat{V}_k(S)}(\psi(z)||^2]  \geq & \frac{1}{m}\sum_{i=1}^k \hat{\lambda}_i - R^2\sqrt{\frac{ \frac{2}{m}\sum_{i=k+1}^m \hat{\lambda}_i\left(\log(\alpha(m))+\log \frac{2 \sqrt{m}}{\delta}\right)}{m}} \\
& -2R^2\frac{\log(\alpha(m))+\log \frac{2 \sqrt{m}}{\delta}}{m} \\
& + \mathbb{E}_{z\sim\mu}[||\psi(z)||^2] - \frac{1}{m} \sum_{i=1}^m ||\psi(z_i)||^2 .
\end{align*}
Finally we apply \citet[Thm 4]{maurer2009empirical} onto the i.i.d variables $1-\frac{\|\psi(z_i)\|^2}{R^2}\in [0;1]$ and we recall that $\psi(z)\|^2= \kappa(z,z)$ for all $z\in\mathcal{Z}$ to conclude that with probability at least $1-\delta$
\[ \mathbb{E}_{z\sim\mu}[||\psi(z)||^2] - \frac{1}{m} \sum_{i=1}^m ||\psi(z_i)||^2 \geq -\sqrt{\frac{\mathbb{V}_m(S)\log \frac{2}{\delta}}{m}} - R^2\frac{7\log(\frac{2}{\delta})}{3(m-1)}.   \]
We have our second bound, with probability $1-2\delta$.

Our last step to conclude the proof is to apply an union bound with $\delta'= \frac{\delta}{2}$.
\end{proof}

We have the following corollary for translation-invariant kernels.

\begin{corollary}
We take the notation of \Cref{th: main_bounds_TS}. Assume we have a translation invariant kernel. If $k$ is big enough such that $\frac{2}{m}\sum_{i=k+1}^m \hat{\lambda}_i \approx \frac{\log(\alpha(m)}{m}$ then we have with probability at least $1-\delta$:

\begin{equation*}
\mathbb{E}_\mu[|| P_{\hat{V}_k(S)}^\bot(\psi(x)||^2]  \leq \frac{1}{m}\sum_{i=k+1}^m \hat{\lambda}_i +O\left(\frac{\log(\alpha(m)}{m}\right),
\end{equation*}
and
\begin{equation*}
\mathbb{E}_\mu[ ||P_{\hat{V}_k(S)}(\psi(x)||^2]  \geq \frac{1}{m}\sum_{i=1}^k \hat{\lambda}_i - O\left(\frac{\log(\alpha(m)}{m}\right).
\end{equation*}
and asymptotically, $\frac{\log(\alpha(m))}{m} \sim |\mathcal{Z}|\frac{\log(m)}{m}$.
\end{corollary}

\begin{proof}
We remark that for translation invariant kernels, one has for all $z_i,z_j, \kappa(z_i,z_i)= \kappa(z_j,z_j)$ thus $\mathbb{V}_m(S)=0$. Finally our additionnal assumption ensure us that
$$\sqrt{\frac{ \frac{2}{m}\sum_{i=k+1}^m \hat{\lambda}_i\left(\log\varepsilon_{m,\delta}\right)}{m}} = O\left(\frac{\log\alpha(m)}{m}\right),$$
since $\log(\frac{4\sqrt{m}}{\delta}) = o\left( \log(\alpha(m)  \right)$. The asymptotic rate is given by \Cref{prop: asymptotic_behavior_alpha}.
\end{proof}

\subsection{Proof of \Cref{th: main_result_2}}

\label{sec: mhameddi for kpca}

We now exploit \citet[Corollary 1]{mhammedi19pac} to obtain \Cref{th: main_result_2}. We recall that $n=m/2$ and we split our $m$-sample $S$ onto two samples $S_1,S_2$ of same size $n$ ($S_1$ contains the $m/2$ first points of $S$, $S_2$ holds the rest).

We define $\mathcal{V}= \cup_{k=1}^m \mathcal{F}_k$ the set of all considered subspaces and recall \Cref{th: main_result_2} below

\begin{theorem}
\label{th: main_bounds_mhameddi}
If we perform PCA in the feature space defined by the Aronszjan mapping $(\psi,\mathcal{H})$ of a kernel $\kappa$, then, for a fixed $k\in [m]$, there exists $\mathcal{P}, \mathcal{A}, \mathcal{C}>0$, such that with probability of at least $1-\delta$ over an $m$-sample $S= (z_1,...,z_m)$, if we project new  data $z$ onto the space $\hat{V}_k(S)$, we have

\begin{equation*}
\mathbb{E}_{z\sim \mu}[||P_{\hat{V}_k^\bot(S)}(\psi(z))||^2]  \leq \frac{1}{m}\sum_{i=k+1}^m \hat{\lambda}_i + \min_{\ell\in[m]} \left[ \mathcal{P} \cdot \sqrt{\frac{R_{m,k,\ell} \cdot\varepsilon_{\delta, m}}{m}} +\mathcal{A} \cdot \frac{\varepsilon_{\delta, m}}{m}
+\mathcal{C} \cdot \sqrt{\frac{R_{m,\ell}^{\prime} \cdot \varepsilon_{\delta, m}}{m}} \right],
\end{equation*}
and
\begin{multline*}
\mathbb{E}_{z\sim \mu}[||P_{\hat{V}_k(S)}(\psi(z)||^2] \\ \geq \frac{1}{m}\sum_{i=1}^k \hat{\lambda}_i - \min_{\ell \in [m]} \left[ \mathcal{P} \cdot \sqrt{\frac{R_{m,k,\ell} \cdot\varepsilon_{\delta, m}}{m}} +\mathcal{A} \cdot \frac{\varepsilon_{\delta, m}}{m}
+\mathcal{C} \cdot \sqrt{\frac{R_{m,\ell}^{\prime} \cdot \varepsilon_{\delta, m}}{m}} \right]
- \sqrt{\frac{\mathbb{V}_m(S)\log \frac{2}{\delta}}{m}} - \frac{7\log(\frac{2}{\delta})}{3(m-1)},
\end{multline*}
where $\varepsilon_{m,\delta}= O\left( \log(\alpha(m)) + \log\left( \frac{\log(m)}{\delta}   \right)\right)$, $\alpha(m)=\binom{|\mathcal{Z}|+m-1}{|\mathcal{Z}|-1}$ and
\begin{align*}
\mathbb{V}_m(S) & = \frac{1}{m(m-1)} \sum_{1\leq i<j\leq m } (\kappa(z_i,z_i)-\kappa(z_j,z_j))^2.
\end{align*}

\end{theorem}

\begin{proof}
Let $k\in[m]$.
Our proof follows the same route than the one of \Cref{th: main_bounds_TS}: to apply  \citep[Corollary 1]{mhammedi19pac}, we define first our deterministic estimator which is the application $\hat{f}: S \rightarrow (\hat{V}_1(S),..,\hat{V}_m(S))$. So we are able to choose among $m$ different predictors when we refer to $\hat{f}(S_1)$ for instance.
Then we choose our two priors to be the same uniform distribution $P$ over $\mathcal{F}_k$ and for posterior $Q$ the Dirac in $\hat{V_k}(S)$. Then we have by \Cref{l: combinatorics_KL}:
\[\KL(Q\|P)=\log(|\mathcal{F}_k|) \leq \log(\alpha(m)).  \]
This choice of posterior ensure several points: first
$$L(G_Q)= \mathbb{E}_{z\sim\mu}[||P_{\hat{V}_k(S)}^\bot(\psi(z))||^2],$$ and
$$L_m(G_Q)= \frac{1}{m}\sum_{i=1}^m||P_{\hat{V}_k}^\bot(\psi(z_i))||^2.$$
Second, thanks to \Cref{p: characterization_eigenvalues}, we know that $\frac{1}{m}\sum_{i=1}^m||P_{\hat{V}_k}^\bot(\psi(z_i))||^2 = \sum_{i=k+1}^m \hat{\lambda}_i$.

Finally, as precised above, given our deterministic estimator, we have to choose $\hat{f}(S_1)$ inside the $m$-tuple $(\hat{V}_1(S_1),..,\hat{V}_m(S_1))$, same for $\hat{f}(S_2)$. Thus \citet[Corollary 1]{mhammedi19pac} gives us with  probability $1-\delta$, for any $\ell\in [m]$:

\begin{multline*}
\mathbb{E}_{z\in\mu}[||P_{\hat{V}_k^\top(S)}(\psi(z)||^2] \\ \leq  \frac{1}{m}\sum_{i=k+1}^m \hat{\lambda}_i(S) + \mathcal{P} \cdot \sqrt{\frac{R_{m,k,l} \cdot\left(\operatorname{COMP}_{m}+\varepsilon_{\delta, m}^\prime\right)}{m}}
+\mathcal{A} \cdot \frac{\operatorname{COMP}_{m}+\varepsilon_{\delta, m}^\prime}{m}
+\mathcal{C} \cdot \sqrt{\frac{R_{m,\ell}^{\prime} \cdot \varepsilon_{\delta, m}^\prime}{m}},
\end{multline*}
where $\operatorname{COMP}_{m} = 2 \mathrm{KL}(Q\|P)$ and $\varepsilon_{\delta, m}^\prime \propto \log\left( \frac{\log(m)}{\delta}   \right)$. Thus: $\operatorname{COMP}_{m}+\varepsilon_{\delta, m}^\prime= \varepsilon_{\delta,m}$
as defined in our theorem. Finally, bounding $\varepsilon_{\delta, m}^\prime$ by $\varepsilon_{\delta, m}$ and taking the minimum over $\ell$ gives us the first bound. We obtain the second bound similarly to the proof of \Cref{th: main_bounds_TS} by applying again \citet[Theorem 4]{maurer2009empirical}.

\end{proof}

\bibliography{biblio.bib}       

\end{document}